\documentclass[11pt]{article}
\usepackage{subcaption}
\usepackage{amsmath,amsfonts,amssymb}
\usepackage{graphicx}
\usepackage{setspace}
\usepackage{csquotes}
\usepackage{hyperref}

\title{A Counterexample in Image Registration}

\author{Serap A.~Savari \\ Texas A\&M University, College Station, TX 77843-3128, USA}

\begin{document}

\maketitle

\begin{abstract}
  Image registration is a widespread problem which applies models about
  image transformation or image similarity to align discrete images of the
  same scene.  Nevertheless, the theoretical limits on its accuracy are not
  understood even in the case of one-dimensional data.  Just as Nyquist's
  sampling theorem states conditions for the perfect reconstruction of signals
  from samples, there are bounds to the quality of reproductions of quantized
  functions from sets of ideal, noiseless samples in the absence of additional
  assumptions.  In this work we estimate spatially-limited piecewise constant
  signals from two or more sets of noiseless sampling patterns.  We mainly
  focus on the energy of the error function and find that the uncertainties
  of the positions of the discontinuity points of the function depend on
  the discontinuity point selected as the reference point of the signal.
  As a consequence, the accuracy of the estimate of the signal depends on the
  reference point of that signal.
\end{abstract}

\section{Introduction}
  The correlation and alignment of images of a scene is a problem investigated
  by the computer vision, photogrammetry and remote sensing communities for
  many years \cite{tong}.
  Moreover, other fields like medical imaging and microscopy apply
  image registration techniques \cite{tong}.
  Therefore, it is essential to thoroughly understand every facet of
  image registration beginning with first principles.

  There is a large literature investigating image registration from either
  the viewpoint of geometric transformations and/or the information
  offered by frequency domain representations \cite{tong}, \cite[\S 2.5]{ip}.
  Our objective here is different.  We focus on a one-dimensional problem
  to raise a question about conventional viewpoints.
  Discete image registration seeks to correlate information based on samples.
  Nyquist's sampling theorem \cite{nyquist} may be the most famous result
  about obtaining information from samples, and it depends on the nature of the signal being sampled. Digital images are generally
  quantized \cite[\S 2.4]{ip}, so the impact of quantization on discrete
  image registration needs investigation.  In \cite{savari} we mainly
  concentrate on the static forward problem of characterizing the possible
  sets of samples of the support of a spatially limited piecewise constant
  function when the sampling interval is $T$ and the pieces of the support
  each have length at least $T$.  We also describe in \cite{savari} how
  the inverse problem of subinterval estimation relates to partitions of a
  unit hypercube.  In this work, we extend \cite{savari} to another important
  inverse problem and see results that may be unexpected.  We continue to
  assume that we have ideal noiseless samples of the support of the function,
  and we consider how to convert sets of sampling patterns into an estimate
  of the function.  We mainly study the energy of the estimation error, but
  we also comment on the absolute error criterion.  We find that the
  uncertainties associated with the positions of the discontinuity points
  of the function generally depend on the discontinuity point selected as
  the reference point of the signal.  Therefore, the accuracy of the estimate
  depends on that point.

  The plan for the rest of the paper is as follows.  In Section~2, we review
  some of the notation and results from \cite{savari}.
  In Section~3, we consider the problem of estimating the support of
  spatially-limited piecewise constant function from ideal and noiseless
  collections of sampling patterns..  In Section~4, we conclude.

\section{Preliminaries}
In \cite{savari} we define our underlying function
\begin{equation}
g(t) \; = \; \left\{ \begin{array}{ll}
g_1, & 0 \leq t < R_1 \\
g_i, & \sum_{j=1}^{i-1} R_j  \leq t < \sum_{j=1}^{i} R_j, \; i \geq 2  \\
0, & \mbox{otherwise};
\end{array}
\right.
\end{equation}
here $g_1 \neq 0, \; g_m \neq 0,$ and $g_i \neq g_{i+1}, i \in \{1, \ \dots ,
\ m-1\}$.  It is convenient to define $g_0 = g_{m+1} = 0 $.
Given the sampling interval $T$, region length $R_i$ satisfies
\begin{displaymath}
R_i = (n_i-f_i)T, i \in \{1, \ \dots , \ m\}, 
\end{displaymath}
where for each $i, \; n_i$ is an integer that is at least two and
$0 < f_i \leq 1$.
The number $\eta_i$ of samples from region $i$ depends on the position
of the sampling grid.  Let $\Delta_i \in [0, \ T)$ denote the
distance from the left endpoint of region $i$ to the first sampling
point within the region.
Let $\lfloor x \rfloor , \; x \geq 0,$ denote the
integer part of $x$.  For non-negative integers $K$, define
\begin{eqnarray*}
  \kappa (i, \ K) & = & \lfloor \sum_{j=0}^{K} f_{i+j}  \rfloor \\
  d_{i,K} & = &  \sum_{j=0}^{K} n_{i+j} -  \kappa (i, \ K) .
\end{eqnarray*}

We apply the following extension of \cite[Proposition 1]{savari}:

\noindent {\bf Proposition 1:} For $i \geq 1$ and $i+K \leq m$, the
cumulative number of samples over regions $i, i+1, \ \dots , \ i+K$ satisfies
\begin{displaymath}
\sum_{j=0}^K \eta_{i+j} \; = \; \left\{ \begin{array}{ll}
  d_{i,K}, & 0 \leq \Delta_i < (1+\kappa (i, \ K) - \sum_{j=0}^{K} f_{i+j})T \\
  d_{i,K}-1, &  (1+\kappa (i, \ K) - \sum_{j=0}^{K} f_{i+j})T \leq \Delta_i 
 < T .
\end{array}
\right.
\end{displaymath}
\noindent {\bf Proof:} If we start counting samples from a sample point at
distance $\Delta_i$ from the left endpoint of region $i$, then the location
of sample $d_{i,K}$ occurs at $\Delta_i + (d_{i,K} -1) T$
while the distance from the left endpoint  of region $i$ to the right
endpoint of region $i+K$ is $\left(d_{i,K} - \left(
\sum_{j=0}^{K} f_{i+j} -  \kappa (i, \ K) \right) \right) T.
\; \Box$

Suppose throughout that there are no integers $i$ and $K$ for which
$\sum_{j=0}^{K} f_{i+j}$ is an integer.  Then
Proposition~1 indicates that we know $\sum_{j=0}^{K} R_{i+j}$ to within $T$
if our data offers two values of $\sum_{j=0}^{K} \eta_{i+j}$ because we then
know $d_{i,K}$.  If our data offers one value of $\sum_{j=0}^{K} \eta_{i+j}$,
then we do not know $d_{i,K}$, and we can only infer
$\sum_{j=0}^{K} R_{i+j}$ to within $2T$.  The introduction of \cite{savari}
elaborates on this point with an example inspired by measurements.

The following is part of \cite[Proposition 4]{savari}:

\noindent {\bf Proposition 2:}
The projection 
$(\eta_i , \ \eta_{i+1} , \ \dots , \ \eta_{i+k-1} ), \; i \geq 1, \
i+k-1 \leq m$, of $(\eta_1 , \ \dots , \ \eta_m )$
is defined $k+1$ intervals containing $\Delta_i$, where $k-1$ of these
intervals remain fixed from the projection
$(\eta_i , \ \eta_{i+1} , \ \dots , \ \eta_{i+k-2} )$,
and the last interval for the latter projection is split into two intervals
for the former projection.

Therefore, the support of $g(t)$ has $m+1$ possible distinct patterns
of sampling counts.  We will also allude to \cite[Theorem 5]{savari}
which describes the unique pattern 
$(\eta_i^{'} , \ \eta_{i+1}^{'} , \ \dots , \ \eta_{i+K-1}^{'} )$
which has  both extensions
$\eta_{i+K} = n_{i+K} $ and $\eta_{i+K} = n_{i+K} -1$
among the $K+2$ possible outcomes for \\
$(\eta_i , \ \eta_{i+1} , \ \dots , \ \eta_{i+K} )$.

\section{Main Results}
The paper \cite{savari} addresses one inverse problem, but there are other
important ones.  We next consider how to estimate $g(t)$ from a collection
of ideal, noiseless sampling paterns of its support.

We define $g(t)$ in \cite{savari} and in (1) with the left endpoint of
the support of the function at zero.  However, while one fixed reference
point is useful in obtaining the results of \cite{savari}, it is arbitrary.
For example, one could alternatively set the right endpoint of the support
of $g(t)$ to zero and process the sampling patterns from right to left.
We consider estimation from an even more general viewpoint.  The function
$g(t)$ has $m+1$ discontinuity points and we will consider the possibility
of setting any of their locations to zero so that we will be working with
translations of the original $g(t)$.  This generally results in estimates
of the various translations which differ in accuracy, and we seek the one
with the lowest energy of estimation error.  We begin by considering the
case where we have all $m+1$ possible observations of sampling counts
of the support and then proceed to the case of fewer observations.

\subsection{Estimation from a Complete Set of Sampling Patterns}
Suppose $g^{(l)} (t), \ i \in \{1, \ \dots , m\}$ is the translation of
  $g(t)$ when discontinuity point $l$ is set to zero, and let $D_i^{(l)}
  , \ i \in \{0, \ \dots , m\}$ denote the position of discontinuity point
    $i$ for this translation.  Then in $R_0=0$,
    \begin{eqnarray*}
      g^{(l)} (t) & = & g(t + \sum_{i=0}^{l} R_i ) \\
      D_i^{(l)} & = &  \left\{ \begin{array}{ll}
        - \sum_{j=i+1}^{l} R_j ,        &  i \in \{0, \ \dots , l-1\}     \\
        0, & i=l \\
         \sum_{j=l+1}^{i} R_j ,        &  i \in \{l+1, \ \dots , m\}     

\end{array}
\right.
      \end{eqnarray*}

    We have the following result.

    \noindent {\bf Lemma 3:} Given all $m+1$ patterns of sampling counts
    for the support of $g(t)$, the discontinuity points 
    $D_i^{(l)}, i \neq l$, are each known to within $T$.

    \noindent {\bf Proof (by induction on the value of $|l-i|$):}
    By Proposition 1, we know that     $D_i^{(l)}, i \neq l$,
    to within $T$ if our observations provide two values of
    \begin{equation}
\left\{ \begin{array}{ll}
         \sum_{j=i+1}^{l} \eta_j ,        &  i \in \{0, \ \dots , l-1\}     \\
         \sum_{j=l+1}^{i} \eta_j ,        &  i \in \{l+1, \ \dots , m\}  .

\end{array}
\right.
    \end{equation}
    By Proposition 1, the result holds at $i \in \{l-1, l+1 \}.$
    For $i-l \geq 2$, Proposition 2 and \cite[Theorem 5]{savari}
    imply that there is a partial observation 
    $(\eta_{l+1}^{'} , \  \dots , \ \eta_{i-1}^{'} )$
    for which $(\eta_{l+1}^{'} , \  \dots , \ \eta_{i-1}^{'}, \ n_i )$
    and $(\eta_{l+1}^{'} , \  \dots , \ \eta_{i-1}^{'}, \ n_i -1 )$
    are both in the collection of sampling patterns.  By the symmetry
    of processing sampling counts from left to right or from right to left,
    if $i-l \leq -2$, then there is a partial observation for which
    $(n_{i+1}, \ \eta_{i+2}^{'} , \  \dots , \ \eta_{l}^{'} )$
    and $(n_{i+1}-1, \ \eta_{i+2}^{'} , \  \dots , \ \eta_{l}^{'} )$
    are both in the collection of sampling count patterns. $\Box$

    Lemma 3 implies that $g^{(l)} (t)$ is known outside of $m$ intervals
    of length $T$ which contain $D_i^{(l)}, i \neq l$.

    Define $C_l^{(l)}=0$ and integers $C_i^{(l)}, i \neq l$, by
    \begin{equation}
\left\{ \begin{array}{ll}
  \sum_{j=i+1}^{l} \eta_j  \in \{C_i^{(l)}-1, \ C_i^{(l)}\},
  &  i \in \{0, \ \dots , l-1\}     \\
  \sum_{j=l+1}^{i} \eta_j \in \{C_i^{(l)}-1, \ C_i^{(l)}\},
  &  i \in \{l+1, \ \dots , m\}  
\end{array}
\right.
    \end{equation}
    for all sampling patterns.  We have the following result.

    \noindent {\bf Lemma 4:} The complete collection of $m+1$ patterns of
    sampling counts determines $g^{(l)} (nT)$ for all integers $n$.

    \noindent {\bf Proof:} By the definition of $C_i^{(l)}, i \neq l$,
    \begin{displaymath}
           - C_i^{(l)} T \; < \; D_i^{(l)} \;       < \; -(C_i^{(l)}-1)T, \;
        i \in \{0, \ \dots , \ l-1\}     
    \end{displaymath}
    \begin{equation}
(C_i^{(l)}-1)T \; < \; D_i^{(l)} \; < \;  C_i^{(l)} T, \;
  i \in \{l+1, \ \dots , m\}  
    \end{equation}

    Observe that $D_{l-1}^{(l)} \in (-n_l T, -(n_l-1) T),$
    and $D_{l+1}^{(l)} \in ((n_{l+1} -1) T, n_{l+1} T).$
    Since $n_i \geq 2$ for all $i$, these intervals do not contain
    $D_{l}^{(l)} =0.$  Next suppose $i<l-1$ and 
    $( \eta_{i+2}^{'} , \  \dots , \ \eta_{l}^{'} )$ is a partial
    sample pattern for which
    \begin{displaymath}
      \eta_{i+2}^{'} +  \dots + \eta_{l}^{'} = C_{i+1}^{(l)} .
    \end{displaymath}
    Then $C_i^{(l)} = C_{i+1}^{(l)} + \eta_{i+1}^{'}
    \geq C_{i+1}^{(l)} + 1$, so the open intervals containing
    $D_i^{(l)}$ and $D_{i+1}^{(l)}$ do not overlap.
    Likewise, if  $i>l+1$ and 
    $( \eta_{l+1}^{'} , \  \dots , \ \eta_{i-1}^{'} )$ is a partial
    sample pattern for which
        \begin{displaymath}
      \eta_{l+1}^{'} +  \dots + \eta_{i-1}^{'} = C_{i-1}^{(l)} ,
    \end{displaymath}
    then $C_i^{(l)} = C_{i+1}^{(l)} + \eta_{i}^{'}
    \geq C_{i-1}^{(l)} + 1$, so the open intervals containing
    $D_i^{(l)}$ and $D_{i+1}^{(l)}$ do not overlap.

    To simplify the subsequent discussion, for $i \neq l$ define
    $G_{i,L}^{(l)}$     and  $G_{i,R}^{(l)}$ so that
    \begin{equation}
      D_i^{(l)} \in (G_{i,L}^{(l)} , \ G_{i,R}^{(l)} ),
          \end{equation}
where $G_{i,L}^{(l)}$     and  $G_{i,R}^{(l)}$ are specified by (4).
Then $g^{(l)} (0) = g_{l+1}$, and for $i \neq l$
$g^{(l)} (G_{i,L}^{(l)}) = g_i$ 
and $g^{(l)} (G_{i,R}^{(l)}) = g_{i+1}$.
Since $g^{(l)} (t)$ is piecewise constant, for
$i \in \{1, \ \dots , m\} , \; g^{(l)} (t) = g_i$ when
$D_{i-1}^{(l)} \leq t < D_i^{(l)} $.  Hence, g(t) is known when $t$ is
an integer multiple of $T. \; \Box$

Let $G_{l,R}^{(l)}=0.$  The main result of this subsection is

\noindent {\bf Theorem 5:} Given all $m+1$ patterns of sampling counts for
$g^{(l)} (t)$, the estimate
\begin{equation}
  \hat{g}^{(l)}(t)
  \; = \; \left\{ \begin{array}{ll}
      0, & t \leq G_{0,L}^{(l)} \; \mbox{or} \; t \geq G_{m,L}^{(l)} \\
g_i, & i \neq l, \; G_{i-1,R}^{(l)}  \leq t \leq G_{i,L}^{(l)} \\
g_1, & G_{l-1,R}^{(l)}  \leq t < 0 \\
\frac{g_i + g_{i+1}}{2}, & i \neq l, \;
G_{i,L}^{(l)}  < t < G_{i,R}^{(l)} 
\end{array}
\right.
\end{equation}
is an extremum of the problem of minimizing the energy of the estimation
error with respect to $  g^{(l)}(t)$ and the corresponding estimation error
is
\begin{displaymath}
  \sum_{i \neq l} \left( \frac{g_i - g_{i+1}}{2} \right)^2 T.
  \end{displaymath}
\noindent {\bf Proof:} By the proof of Lemma 4, $g^{(l)} (t)$ is known
except for the $m$ open intervals containing the discontinuity points
$D_i^{(l)}, \; i \neq l$, so we only need to estimate
$g^{(l)} (t)$ in these intervals.  Therefore, for the class of estimates
of interest, the energy $\epsilon$ of the estimation error is
\begin{equation}
  \epsilon = \sum_{i \neq l}  \left( \int_{G_{i,L}^{(l)}}^{D_i^{(l)}}
  (\hat{g}^{(l)} (t) - g_i)^2 \ dt
  + \int_{D_i^{(l)}}^{G_{i,R}^{(l)}} (\hat{g}^{(l)} (t) - g_{i+1})^2 \ dt
  \right)
\end{equation}

We study how the preceding expression varies with $R_k, \;
k \in \{1, \dots , m \}$.  For $l<m$ we begin with $R_m$, which appears
only in ${D_i^{(l)}}$ and subsequently work backward to $R_{l+1}$.
We have
\begin{displaymath}
  \frac{\partial \epsilon}{\partial R_m}  =
  (\hat{g}^{(l)} ({D_m^{(l)}}  ) - g_m)^2 
  - (\hat{g}^{(l)} ({D_m^{(l)}}   ) - g_{m+1})^2 .
\end{displaymath}
In the absence of other information ${D_m^{(l)}}$ can fall anywhere in \\
$({C_m^{(l)}}-1)T, {C_m^{(l)}}T)$.
Therefore, an extremum for this optimization problem satisfies
$  \hat{g}^{(l)} (t) = \frac{g_m + g_{m+1}}{2} = \frac{g_m}{2}, \; l<m,
\; t \in ({C_m^{(l)}}-1)T, {C_m^{(l)}}T)$, and the energy of the error
in this interval is $\left( \frac{g_m - g_{m+1}}{2} \right)^2 T$,
independent of
${D_m^{(l)}}$.  Therefore, for $l<m$ we have reduced the problem to the
study of the error associated with the remaining $m-1$ discontinuity points,
and for $l<m-1, \; R_{m-1}$ remains only in ${D_{m-1}^{(l)}}$.
For $l>0, \; R_1$ appears only in ${D_{0}^{(l)}}$, and by similar reasoning
an extremum in this case satisfies 
$  \hat{g}^{(l)} (t) = \frac{g_0 + g_{1}}{2} = \frac{g_1}{2}, \; l>0,
\; t \in -(C_0^{(l)}T, -({C_0^{(l)}}-1)T)$ with an energy of error
$\left( \frac{g_0 - g_{1}}{2} \right)^2T$ in that interval, independent of
${D_0^{(l)}}$.  For $l>0$ we have again reduced the problem by one interval.
The argument follows by successively removing uncertain intervals. $\Box$

Observe that the value of $l$ for which the energy of the estimation error
between $g^{(l)} (t)$  and $\hat{g}^{(l)} (t)$ is minimized is
$\arg \max_k | g_k - g_{k+1} |$, so large discontinuities impact accuracy.
Note that if $\hat{g}^{(l)} (t)$ is constrained to be a constant
$\hat{g}_i$ for the interval containing ${D_i^{(l)}}, i \neq l$, then the
absolute error associated with $\hat{g}^{(l)} (t)$ is at most
\begin{displaymath}
  \sum_{i \neq l} T \max \{ | \hat{g}_i - g_i |, \; | \hat{g}_i - g_{i+1} |\};
\end{displaymath}
this bound is minimized when $\hat{g}_i = \frac{g_i + g_{i+1}}{2}, \;
  i \neq l$.

  \subsection{Estimation from a Partial Set of Sampling Patterns}

  Theorem 5 has established that even with the presence of a full set of
  sampling patterns the accuracy of minimum error energy estimation depends
  on the reference point of the signal.  However, the following small example
  hints at why the problem becomes even more interesting in the presence
  of a partial set of sampling patterns.

  \noindent {\bf Example 6:} Suppose $m=3$ and we have pattern sampling
  counts \\ $(\eta_1 , \ \eta_2 , \ n_3)$ and $(\eta_1 , \ \eta_2 , \ n_3-1)$
  for some integers $\eta_1 \geq 2$ and $\eta_2 \geq 1.$
  If $l=0$, then ${D_0^{(0)}}=0$,  the locations of
  ${D_1^{(0)}}$ and ${D_2^{(0)}}$ each have uncertainty $2T$, and
  the location of ${D_3^{(0)}}$ has uncertainty $T$.
  If $l=3$, then ${D_3^{(3)}}=0$, and the locations of
  ${D_0^{(3)}}, {D_1^{(3)}}$ 
  and ${D_2^{(3)}}$ each have uncertainty $T$.

  In a longer version of this paper we will go through the derivation of a
  generalization to Theorem 5.  Here we merely introduce the notation and
  the result.  Let $U \subseteq \{0, \ \dots , m \} \setminus \{l \}$
  be the collection of indices for which the location of
  ${D_i^{(l)}}$ is not known to
  within $T$ when $i \in U$, and let $U^c = 
  \{0, \ \dots , m \} \setminus U \cup \{l \}$.  $U$ and $U^c$ depend on $l$,
  but we use simpler notation for them and for the coming definitions.
  Define ${C_l^{(l)}}=0$, and for $i \in U^c$ define ${C_i^{(l)}}$ by (3).
  For $i \in U$ define ${C_i^{(l)}}$ to be the single value of the sum
  associated with (2).  Let $S^{+}$ denote the set of indices $i$ with
  $i \geq l+1, \; i \in U, \; i+1 \in U, \; \eta_i^{'} > 1$ for
  some observation and $\eta_{i+1} = 1$ for all observations.
Let $S^{-}$ denote the set of indices $i$ with
  $i \leq l-1, \; i \in U, \; i-1 \in U, \; \eta_{i+1}^{'} > 1$ for
some observation and $\eta_{i} = 1$ for all observations.
Let $s^{+} = | S^{+} |$ and $s^{-} = | S^{-} |$.
Let $t_1^{+} < \dots < t_{s^{+}}^{+} $ be the elements of $S^{+}$ and
$t_1^{-} > \dots > t_{s^{-}}^{-} $ be the elements of $S^{-}$.
For $j \in \{1, \dots , s^{+} \}$, let $\lambda_j^{+} \geq 1$
be the integer for which $\eta_{t_j^{+} + \lambda_j^{+}} = 1$ and
either ${t_{s^{+}}^{+} + \lambda_{s^{+}}^{+}} = m$ or
$\eta_{t_j^{+} + \lambda_j^{+} +1}^{'} > 1$ for some observation.
For $j \in \{1, \dots , s^{-} \}$, let $\lambda_j^{-} \geq 1$
be the integer for which $\eta_{t_j^{-} + 1- \lambda_j^{-}} = 1$ and
either ${t_{s^{-}}^{-} + 1-\lambda_{s^{-}}^{-}} = 1$ or
$\eta_{t_j^{-} - \lambda_j^{-} }^{'} > 1$ for some observation.
For $j \in \{ 1, \dots , s^{+} \},$ let
$V_j^{+} = \{ t_j^{+} , \dots , t_j^{+} + \lambda_j^{+} \} \subset U.$
For $j \in \{ 1, \dots , s^{-} \},$ let
$V_j^{-} = \{ t_j^{-} - \lambda_j^{-} , \dots , t_j^{-} \} \subset U.$
Define
\begin{displaymath}
  V = U \setminus \left( \left( \bigcup_{j=1}^{s^{+}} V_j^{+} \right)
    \cup \left( \bigcup_{j=1}^{s^{-}} V_j^{-} \right) \right) .
\end{displaymath}
For $j \in \{1, \dots , s^{+} \}$, let
\begin{displaymath}
  B_j^{+} = \frac{G_{t_j^{+} + \lambda_j^{+}, R}^{(l)} -
    G_{t_j^{+} , L}^{(l)}}{T} - 1;
\end{displaymath}
for $j \in \{1, \dots , s^{-} \}$, let
\begin{displaymath}
  B_j^{-} = \frac{G_{t_j^{-} , R}^{(l)} - 
    G_{t_j^{-} - \lambda_j^{-}, L}^{(l)}}{T} - 1;
\end{displaymath}

Then we have

\noindent {\bf Theorem 7:} Suppose the estimate
$\hat{g}^{(l)} (t)$ of the function $g^{(l)} (t)$ is piecewise constant and
satisfies
\begin{displaymath}
  \hat{g}^{(l)}(t)
  \; = \; \left\{ \begin{array}{ll}
    \gamma_n, & (n-1)T <t < nT, \; n \in \{
    \frac{G_{0,L}^{(l)}}{T} +1, \dots ,     \frac{G_{m,R}^{(l)}}{T} \} \\
      0, & \mbox{otherwise}
\end{array}
\right.
\end{displaymath}
for some constants $\{ \gamma_n \}$.
Then $\hat{g}^{(l)} (t)$ will optimize the worst case contribution to the
energy of the estimation error if
\begin{displaymath}
  \hat{g}^{(l)}(t)
  \; = \; \left\{ \begin{array}{ll}
      0, & t \leq G_{0,L}^{(l)} \; \mbox{or} \; t \geq G_{m,L}^{(l)} \\
      g_i, & G_{i-1,R}^{(l)}  \leq t \leq G_{i,L}^{(l)}, \\
      \mbox{ } & i-1 \in U^c \cup V \cup
      \left( \bigcup_{j=1}^{s^{+}} t_j^{+} + \lambda_j^{+} \right)
      \cup \left( \bigcup_{j=1}^{s^{-}} t_j^{-} \right), \\
      \mbox{ } & i \in U^c \cup V \cup
      \left( \bigcup_{j=1}^{s^{+}} t_j^{+} \right)
      \cup \left( \bigcup_{j=1}^{s^{-}} t_j^{-} - \lambda_j^{-} \right) \\
g_1, & G_{l-1,R}^{(l)}  \leq t < 0 
      \end{array}
\right.
\end{displaymath}
or it is
$\frac{g_i + g_{i+1}}{2}$ when
\begin{itemize}
\item $G_{i,L}^{(l)}  < t < G_{i,R}^{(l)} , i \in U^c \cup V $
  \item $G_{i,L}^{(l)}  < t < G_{i,L}^{(l)} + T, i \in
      \cup \left( \bigcup_{j=1}^{s^{+}} t_j^{+} \right)
      \cup \left( \bigcup_{j=1}^{s^{-}} t_j^{-} - \lambda_j^{-} \right)$
      \item
      $G_{i,R}^{(l)}-T  < t < G_{i,R}^{(l)} , i \in
\left( \bigcup_{j=1}^{s^{+}} t_j^{+} + \lambda_j^{+} \right)
      \cup \left( \bigcup_{j=1}^{s^{-}} t_j^{-} \right) $
\end{itemize}
or it is
\begin{displaymath}
  \frac{ \min \{g_{t_j^{+}+i-1} , g_{t_j^{+}+i} , g_{t_j^{+}+i+1} +
    \max \{g_{t_j^{+}+i-1} , g_{t_j^{+}+i} , g_{t_j^{+}+i+1} }{2},
\end{displaymath}
when $G_{t_j^{+},L} + (i-1)T < t < G_{t_j^{+},L} + iT , 
\;  i \in \{ 2, \dots , B_j^{+} \}$
or it is
\begin{displaymath}
            \frac{ \min \{g_{t_j^{-}-(i-1)} , g_{t_j^{-}-i} , g_{t_j^{-}-(i+1)} +
        \max \{g_{t_j^{-}-(i-1)} , g_{t_j^{-}-i} , g_{t_j^{-}-(i+1)} }{2}
\end{displaymath}
when $G_{t_j^{-},R} -iT < t < G_{t_j^{-},R} - (i-1)T , 
\;  i \in \{ 2, \dots , B_j^{-} \}$.
When $U=V$ the corresponding error of the energy function is known to be
\begin{displaymath}
  \sum_{i \in U^c} \left( \frac{g_i - g_{i+1}}{2} \right)^2 T
  +  \sum_{i \in V} \left( \frac{g_i - g_{i+1}}{2} \right)^2 (2T).
  \end{displaymath}

\section{Conclusions} 
Nyquist's sampling theorem indicates that the ability to reproduce a
function from samples and interpolation depends on properties of that
function.  Since discrete image registration seeks to correlate information
about quantized functions from samples, it is important to better understand
the influence of sampling and quantization on this problem.  We focus on a
one-dimensional problem and report an unusual observation:  in the absence
of additional assumptions the uncertainties in the locations of the
discontinuity points of a piecewise continuous function from ideal and
noiseless sets of sampling patterns depend on the vantage point selected
within the signal.  Hence, the accuracy of estimates of such functions
depend on the chosen reference point.

\end{document}